\documentclass[times,twocolumn,final, nopreprintline]{elsarticle}

\pdfoutput=1
\usepackage{placeins}

\RequirePackage{geometry}
\usepackage{framed,multirow}

\usepackage{amssymb}
\usepackage{latexsym}

\usepackage{url}
\usepackage{xcolor}

\usepackage{hyperref}
\usepackage{tabularx}

\usepackage{textcomp}

\definecolor{marked}{rgb}{.349,.8,.1}
\definecolor{marked}{rgb}{.8,.1,.1}
\definecolor{marked}{rgb}{0, 0, 0}

\begin{document}

\begin{frontmatter}

\title{Learning image representations for anomaly detection:  application to discovery of histological alterations in drug development}%

\author[1]{Igor Zingman\corref{cor1}}
\ead{igor.zingman@boehringer-ingelheim.com}
  
\author[2]{Birgit Stierstorfer}
\author[1]{Charlotte Lempp}

\author[1]{Fabian Heinemann\corref{cor1}}
\ead{fabian.heinemann@boehringer-ingelheim.com}
\cortext[cor1]{Corresponding authors}

\address[1]{Drug Discovery Sciences, Boehringer Ingelheim Pharma GmbH and Co., Biberach an der Riß, Germany}
\address[2]{Non-Clinical Drug Safety, Boehringer Ingelheim Pharma GmbH and Co., Biberach an der Riß, Germany}

\begin{abstract}
We present a system for anomaly detection in histopathological images.  In histology, normal samples are usually abundant, whereas anomalous (pathological) cases are scarce or not available. Under such settings, one-class classifiers trained on healthy data can detect out-of-distribution anomalous samples. Such approaches combined with pre-trained Convolutional Neural Network (CNN) representations of images were previously employed for anomaly detection (AD). However, pre-trained off-the-shelf CNN representations may not be sensitive to abnormal conditions in tissues, while natural variations of healthy tissue may result in distant representations. To adapt representations to relevant details in healthy tissue we propose training a CNN on an auxiliary task that discriminates healthy tissue of different species, organs, and staining reagents. Almost no additional labeling workload is required, since healthy samples come automatically with aforementioned labels. During training we enforce compact image representations with a center-loss term, which further improves representations for AD. The proposed system outperforms established AD methods on a published dataset of liver anomalies. Moreover, it provided comparable results to conventional methods specifically tailored for quantification of liver anomalies. We show that our approach can be used for toxicity assessment of candidate drugs at early development stages and thereby may reduce expensive late-stage drug attrition.
  
\end{abstract}

\end{frontmatter}


\section{Introduction}
\label{sec1}
Anomaly detection is the problem of identifying observations that substantially deviate from the expected distribution of normal data in a particular domain. It can be considered as a binary classification task  where examples of one class (anomalies) are not available or rare, whereas the other class (normal) is well represented with a sufficient amount of samples. Since anomalous samples are not available, powerful supervised  classifiers cannot be used. Instead, so called one-class classifiers are employed. They can be either discriminative, which try to find compact support for normal data, e.g the one-class Support Vector Machine (SVM) \citep{ScholkopfPSSW01} and Support Vector Data Description (SVDD) \citep{TaxD04}, or generative, which model the distribution of normal data \citep{abs-1909-11786, RippelMM20}. Outlier samples that are outside of the support or have low probability according to the modeled distribution are detected as anomalies. Anomaly detectors have numerous applications. In the computer vision domain AD was used in industrial inspection for detection of manufacturing defects \citep{roth2022towards, BergmannFSS20}, in surveillance for detection of suspicious behavior in videos \citep{SabokrouFFMK18}, and in healthcare for detection of pathologies in medical images \citep{FernandoGDSF22, ShvetsovaBFSD21, ZEHNDER2022100102}. 

In this paper we approach anomaly detection in histological images captured and digitized with an optical microscope (typically so-called whole slide scanners). Such images are obtained from stained tissue samples, which allows visualization of the different tissue structures. Changes in distribution and morphology of the structures help pathologists to identify abnormal conditions. In the domain of histology, as in many other domains in medicine, images of healthy tissue might be available in abundance\footnote{Particularly, in pre-clinical studies large amounts of healthy tissue from animal models are frequently collected.}, while images with abnormal conditions can be hard to gather, scarce or even be completely absent\footnote{Even in clinical settings, where a biopsy is typically taken to verify disease, comprehensive data for many relevant types of tissue anomalies may still be unavailable.}. Moreover, some abnormal conditions might have never been registered before and therefore are unknown. Under such conditions with a lack of labeled pathological data supervised classification methods are not suited, instead, AD methods can be employed.    

Deep learning algorithms and particularly Convolutional Neural Networks (CNN) became most successful approaches in numerous computer vision tasks, including the domain of computational pathology \citep{SrinidhiCM21} and the AD challenge. 
Features generated by pre-trained CNNs  (also called embeddings or representations) were successfully used in conjunction with one-class classification for AD tasks \citep{abs-1909-11786, RippelMM20, defard2021padim, abs-2002-10445, roth2022towards}. The CNNs are usually pre-trained on large datasets of natural images such as ImageNet \citep{DengDSLL009}. Such feature representations might be less effective in healthcare related tasks that deal with images coming from very different imaging modalities. ImageNet pre-trained CNN representations of histological images might be sensitive to normal variations of tissue structures or staining procedures and on the other hand be not sensitive to abnormal changes in tissue patterns. Therefore, \cite{hoefling2021histonet} hypothesized that adaptation of CNNs to the specific domain of histopathology, may benefit various applications that make use of deep CNN image representations.
In \cite{riasatian2021fine} it was demonstrated that an SVM classifier fed with DenseNet CNN \citep{HuangLMW17} image features that was fine-tuned on a large dataset of histopathological images improves its classification performance.
In \cite{KoohbananiUKKR21} the authors show that a specially designed self-supervision that use unlabeled histopathological images helps to achieve better performance for  classification of data (from the same domain) with limited amount of labels. The gain is due to better feature representations generated by a backbone CNN.

\begin{figure*}[!htb]
\centering
\includegraphics[scale=1]{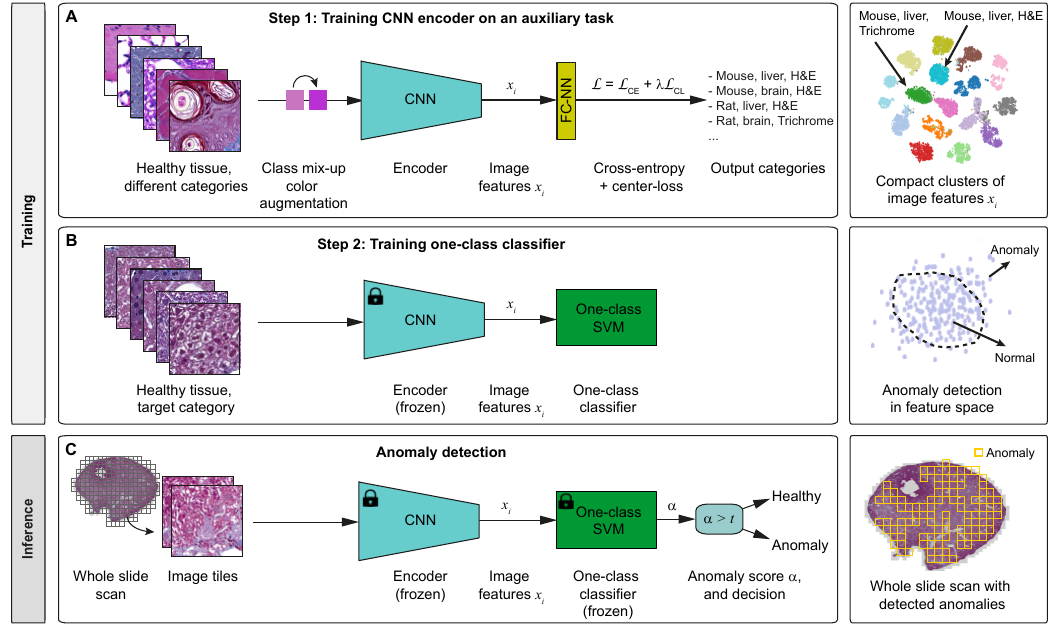}
\caption{\label{Fig:1} Anomaly detection approach. A: Learning image representations with an auxiliary supervised classification task on set of tissue categories. The figure on the right shows t-SNE plot of image features after the training. Clusters marked with arrows correspond to target classes for the anomaly detection task in our experiments. Each color corresponds to a particular category (a combination of specie, organ, and staining). B: Training one-class classifier on a tissue of a category of interest. C: Anomaly detection in tissue of the chosen (target) category. Anomaly score $\alpha$ can be thresholded to output a binary decision.}
\end{figure*}

In our work we present a \textcolor{marked}{recipe} for training a system for the detection of abnormal tiles from Whole Slide Images (WSI) of tissue samples. Particularly, we aim at the detection of anomalies in mouse liver tissue. We use an approach consisting of a CNN-based feature generator, which outputs image representations, followed by one-class SVM classifier, see Fig.~\ref{Fig:1}. We propose techniques listed below to improve image representations for histopathological data and study their influence to the overall performance of the AD system. 

\begin{enumerate}
\item \textbf{Auxiliary supervised classification task:} Since anomaly examples are not available or unknown we cannot train a classifier to distinguish anomalies from normal samples. However, we gathered a large number of healthy tissue samples from different organs of two animal species, mouse and rat, which were stained with two different procedures, hematoxylin \& eosin (H\&E) staining and Masson's trichrome (MT) staining. We define an auxiliary supervised classification task that learns to recognize tissue samples belonging to one of the combinations of organ, specie, and stain that was mentioned above. The target classes of healthy mouse liver tissue stained with H\&E and Masson's Trichrome are also included. The classifier is built from a CNN pre-trained on ImageNet followed by fully connected neural network (FC-NN) layers (see Fig.~\ref{Fig:1} A).  We expect that CNN representations trained on the auxiliary task  will better suit the final task of AD in images of mouse liver tissue, since AD and auxiliary task share the same image domain.  Note that gathering the labels for our auxiliary task present almost no effort.  
\item \textbf{Compact feature representations:} While training a supervised classifier on an auxiliary task we force compactness of representations in the feature space for the normal class of interest (healthy liver mouse), which is thought to be beneficial for later one-class classification. For this purpose we use a center-loss term in the objective function \citep{WenZL016}.
\item \textbf{Class mix-up color augmentation:}  Tissue samples collected for the auxiliary classification task were prepared at different times under slightly different conditions of staining and image acquisition. This may result in trained classifier that is sensitive to differences in stain concentrations or acquisition settings instead of differences in inherent tissue structures. To prevent this unwanted effect we randomly transfer color distributions between tissue classes used for training. We will call this procedure class mix-up color augmentation. Additionally, we apply standard augmentation procedure, where we randomly vary image saturation, hue, brightness, and contrast.
\end{enumerate} 
Once the supervised classifier is trained on an auxiliary task, we use its CNN backbone for generation of image representations and train a one-class classifier on a target class of healthy mouse liver tissue (see Fig.~\ref{Fig:1} B). In our paper we used a one-class SVM \citep{ScholkopfPSSW01}, which was configured in a way that its negative output scores correspond to anomalous detections outside of the estimated normal data support.   

To evaluate the performance of our AD approach, we collected mice liver tissue samples exhibiting the features of Non-Alcoholic Fatty Liver Disease  (NAFLD). We made this dataset along with a set of healthy liver tissue publicly available \citep{dataset}. The published tissue samples were stained with both H\&E and Masson's Trichrome and may be used as a benchmark dataset for anomaly detection in histopathological tissue. We provide a comparison with a sequence of state of the art methods for anomaly detection  \citep{AkcayAB18, roth2022towards,abs-1909-11786,defard2021padim,WangHD021,GudovskiyIK22} that were implemented as a part of Anomalib benchmark \citep{abs-2202-08341web} \textcolor{marked}{and with an anomaly detector developed by \cite{ShvetsovaBFSD21} that has shown a strong performance on digital pathology images.}
In addition, we compare the quality of our feature representations with the features obtained with self-supervised learning (SSL) methods \citep{voigt2023investigation}, both contrastive \citep{chen2020simclr} and non-contrastive \citep{chen2021exploring}.
We also provide an ablation study to show how different parts of our system influence the AD performance. 
The comparative evaluation has shown that our approach outperforms all the other tested methods. 
Moreover, we show that our AD approach behaves similarly to approaches tailored for NAFLD quantification, an immunohistochemical staining and pathologist's assessment using NAFLD activity score (NAS). Recently, also deep leaning supervised classification methods were developed for NAS prediction \citep{heinemann2019deep}.

The underlying motivation for this work is the ability to identify toxicological effects of candidate drugs at early pre-clinical development stages. Toxicological findings are a leading cause of high rates of drug attrition at pre-clinical phases, which is a major cost driver in drug development \citep{waring2015analysis}. We, therefore, test our AD system on a known case of a drug with  known toxicological effects and show that we were able to detect and quantify tissue anomalies (degeneration, necrosis and vacuolation in liver tissue) caused by increasing dose of the drug.  

To summarize, our paper makes the following major contributions:
\begin{itemize}
\item We introduced a method for anomaly detection in tissue, which is mainly based on three techniques for learning effective feature representations of histopathological images: an auxiliary supervised classification task, compact feature representations, and class mix-up color augmentation. 
\item We collected and published a dataset for the evaluation of anomaly detection in histopathological images. The dataset was used to study the importance of each of the proposed techniques and to make performance comparison of our approach with the established and recent AD methods.
\item We have shown that the anomaly detection approach may support the identification of toxicological effects of drug candidates and thereby potentially reduce expensive late-stage drug attrition.
\end{itemize}

\section{Related work}
AD methods mentioned in the previous section consist of two stages, feature generation and one-class classification. Such two-stage design allows application of successful approaches for training feature representations, e.g self-supervised learning \citep{SohnLYJP21} or transfer learning with abundant data sources not directly related to the task at hand \citep{perera2019learning}. 
Learning feature representation usually does not involve an optimization of the AD objective. However, \cite{perera2019learning} suggested to use compactness loss to reduce variance of normal data in the feature space, which may increase the space of possible anomalies that can be detected. 
Although attempts were made to combine the two stages such that a single training procedure is required that directly optimizes a single AD loss function \citep{RuffGDSVBMK18, OzaP19}, two stage methods deem to be more powerful for AD \citep{SohnLYJP21}. 

In our work we adopted several ideas described in \cite{perera2019learning} and applied them to the domain of histopathology with an application in drug discovery. Similarly to \cite{perera2019learning}, we tune image representations using an auxiliary dataset and a compactness loss. In our case the dataset is built from healthy tissue of different organs and species. The difference is that \cite{perera2019learning} used target (one-class) set to optimize compactness loss and an external (multi-class) ImageNet set to optimize descriptiveness loss (cross-entropy). In this case, image representations do not directly learn to distinguish the target class from the rest of the (auxiliary) data. In contrast, we use  a single dataset of normal histological data (without pathological alterations) that includes the target class (mouse liver). This enforces image representations to distinguish the target class from other tissue categories and simultaneously to be compact for the target as well as auxiliary classes. We also suppose that the gain from tuning image representations in our case is larger, since pre-trained CNNs are re-purposed from ImageNet to histopathological images, whereas in \cite{perera2019learning} a pre-trained CNN  was forced to only keep discrimination ability of features for ImageNet categories while improving their compactness for the target class. Another subtle difference to \cite{perera2019learning} is that we used an objective function with a center-loss term that is more similar to the one proposed in \cite{WenZL016}, \textcolor{marked}{which allowed us to enforce compactness for arbitrary classes of tissue during training on the auxiliary dataset, see Sec.~\ref{Sec:ablation_study}.}


One-class classification approaches detect anomalies in a feature (latent) space. An alternative approach is based on image reconstruction methods, such as auto-encoders, where anomaly is detected in an image space by means of measuring a reconstruction error, the deviation of a reconstructed image from an initial one. Since these methods learn to generate images based on a dataset of normal images, anomalies are supposed to have high reconstruction errors. 
Recently, with the development variational auto-encoders \citep{KingmaW13} and generative adversarial networks \citep{RadfordMC15} image reconstruction based methods became more popular for anomaly detection, see for example \cite{SchleglSWLS19, AkcayAB18}. They, however, usually require large amount of data, are harder to train, and to control their behavior. For example, these methods might be prone to reconstruct anomalous regions in an image with a similar success as normal regions. On the other hand, they are known to lose fine details in reconstructed images, the details that may be characteristic to anomalies of an interest. Additionally,  reconstruction models tend to learn low-level image statistics \citep{NalisnickMTGL19}, perhaps, due to used pixel-based reconstruction loss. Advances in development of the reconstruction-based methods gradually improve their weak points, see for example \citep{ShvetsovaBFSD21}.  Until now, however, the best performing AD approaches are based on an outlier detection in a feature space \citep{roth2022towards}. Though some benchmarks \citep{abs-2202-08341web} support this general statement, a comparative evaluation of AD performance is very dependent on the used dataset \citep{ruff2021unifying} and particularly on the types of provided anomalies. Larger, diverse, and more challenging AD datasets are required to reliably rank AD methods and thereby direct research in the field. 

\textcolor{marked}{Some AD methods were specifically tailored and evaluated on histopathological images \citep{ShvetsovaBFSD21, ZEHNDER2022100102, pocevivciute2021unsupervised}. In these works generative methods (Generative Adversarial Networks and Autoencoders) were used to leverage an image reconstruction error within the anomaly score. All these works emphasized disadvantages of using pixel-to-pixel based reconstruction errors in loss functions. Instead, a perceptual loss was used, which is a high-dimensional representation of an image generated by a neural network. Additionally, an anomaly score incorporated image edge-based error in \cite{pocevivciute2021unsupervised}, and SSIM-based \citep{1284395} reconstruction error in \cite{ZEHNDER2022100102} were used, while \cite{ShvetsovaBFSD21} used solely the perceptual loss for anomaly score. One-class classification-based AD approaches by design use (perceptual) image representations and  altogether avoid using image reconstruction-based algorithms. In our work we adapt these representations to specifically capture most important image information for assessing histopathological images. }

\section{Anomaly detection in histological images}

\subsection{AD system overview}
We design an anomaly detection system that is a tile-based processing system. First, whole slide images (WSI) are cut into tiles ($256 \times 256$ pix., 0.44 \textmu m/{pix.}  in our experiments in Sec.~\ref{Sec:Experiments}). Each tile is then processed with the system generating an anomaly score, see Fig. ~\ref{Fig:1} C. We design an AD system that consists of two blocks. The first, the CNN encoder, which encodes an input image tile with a feature vector representation and, the second, one-class classifier that generates an anomaly score $\alpha$ for every feature vector (see Fig.~\ref{Fig:1}).
All anomaly scores from a WSI can then be integrated into a single WSI anomaly score. 

To aggregate tile scores into a single WSI level score we use a simple strategy. We threshold each tile score to decide whether it is anomalous and compute the fraction of anomalous tiles in the WSI. The system is trained in a way (namely, the one-class classifier, see Fig.~\ref{Fig:1} B) that the abnormality of a tile is determined with a threshold of zero.   Other simple aggregation strategies are possible. For example, tile scores can first pass through logistic function with a particular growth coefficient and then summed up to a single WSI score. These simple methods do not take into account spatial correlation between the tiles, which can carry additional diagnostic information. This topic, however, is beyond the scope of this paper. A reader interested in patch (tile) aggregation strategies is referred to \citet{SrinidhiCM21, ilse2018attention, kraus2016classifying}.
 
\subsubsection{CNN encoder: learning image representations}
 \label{Sec:LearningRepresentations}
The critical part of the AD system is a CNN encoder that should generate feature vectors with high representation power allowing discrimination between anomalous and normal tissue structures. On the other, representation vectors should be short enough to generalize to unseen data. Preferably, to facilitate operation of a one-class classifier, representations should be invariant to variations in image appearance due to differences in staining procedure and image acquisition. CNNs that were pre-trained on ImageNet dataset may generate feature vectors that do not have the mentioned above  properties. We therefore propose a few techniques for training CNN to adapt image representations to histopathological images.
\subsubsection{Auxiliary tissue classification task}
 \label{Sec:Auxiliary}
We define an auxiliary classification task, which is depicted in the top of Fig.~\ref{Fig:1} A, with 16 tissue categories, each of which corresponds to a combination of a particular specie, organ, and staining protocol. This incorporates mouse and rat species, H\&E and Masson's Trichrome staining protocols, and \textit{liver, brain, kidney, heart, lung, pancreas, spleen} organs. 
Rat tissue was available only for liver, so that not all combinations of animal specie, organ, and staining were used. Note that no special effort is needed to gather labels (specie, organ, staining) for our training dataset of healthy tissue, since labels come automatically with images.

Instead of defining multi-class classification task, we could use a multi-task learning approach \citep{ThungW18} and define three separate tasks for detection of a specie, organ, and staining type. In that case we would need to define an objective function composed of three weighted terms, with best weights to be found empirically.
To effectively use such a multi-task approach, it would also require gathering images of rat tissue for different organs in addition to liver. We, however, did not experiment with such an approach in this paper and leave this for future work. 

We train a fully convolutional neural network and followed fully connected Neural Network (NN) classifier on the auxiliary task with $256\times 256$ tissue tiles. 1-D feature vectors were obtained with global averaging of the last $8\times 8$ activation map of the fully convolutional network. This allows potential usage of input images of any size.

Instead of global average pooling, an additional convolutional layer could be used that generates a final activation map of $1\times 1$ size. However, this would imply unequal contribution of entries from the last activation map (each of which corresponds to a small patch in the original image). This in turn would contradict the underlying textural structure of histopathological images. In contrast to natural images with objects, all positions in a histological image are semantically equal.

\begin{figure*}[!htb]
\centering
\includegraphics[scale=1]{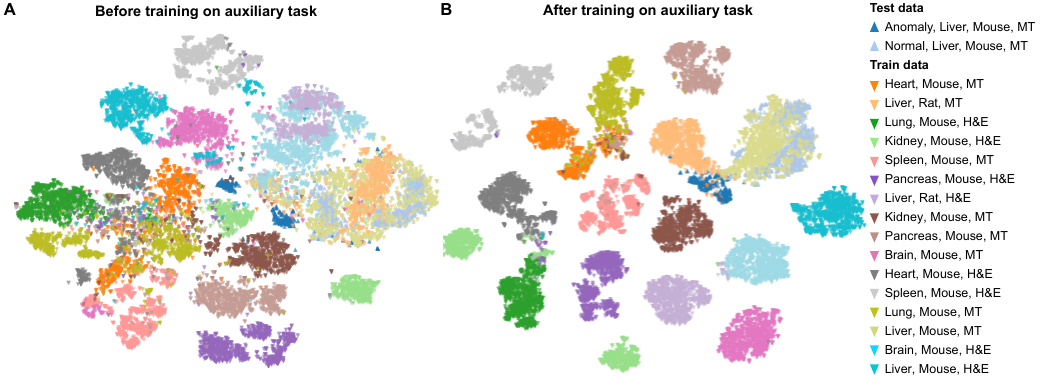}
\caption{\label{Fig:2} t-SNE visualization of feature representations of images \textcolor{marked}{before (A) and after (B)} CNN was trained on the auxiliary task (see Sec.~ \Ref{Sec:Auxiliary}). First two markers in the legend correspond to the test data we use in experiments in the Sec.~ \Ref{Sec:Experiments}, while all the other correspond to training data (healthy tissue from different organs, species, staining) used for the auxiliary task. }
\end{figure*}

We expect that after training on the auxiliary task, the CNN will be able to provide qualitative image representations that can discriminate between histological structures. We also expect that learned representations are to some degree insensitive to variations in appearance of histopathological images due to variations in staining and image acquisition settings. Fig.~\ref{Fig:2} shows a t-SNE visualization \citep{van2008visualizing} of image representations of the training data from our auxiliary task and test data from anomaly detection task within a single plot. \textcolor{marked}{Fig.~\ref{Fig:2}A and Fig.~\ref{Fig:2}B show the representations before and after training the CNN on the auxiliary task. Put attention that most of  anomalous Liver-Mouse-MT test data, marked in dark blue, fall outside of normal Liver-Mouse-MT test and train data, marked in light blue and light olive colors, respectively}. However, anomalous data is much closer to normal data than the other categories, except for Liver-Rat-MT category (light orange). This shows that, as expected, Rat and Mouse tissue of the same organ are much more similar to each other than tissues of different organs.

\subsubsection{Architecture of the neural network}

After experimenting with architectures for CNN encoder we have chosen to use EfficientNet-B0 deep neural network \citep{TanL19} (pre-trained on ImageNet), where we remove the last $9^{th}$ stage with $1\times1$ convolutional layer, such that the number of the output channels $d=320$. After global averaging of $d \; 8\times 8$ activation maps, the CNN encoder outputs a 320-dimensional feature vector, which is fed to the FC-NN classifier as shown in top of Fig.~\ref{Fig:1}. In Sec.~ \ref{Sec:evaluationNAFLD} we will compare different CNN architectures within our AD system, including EfficientNet-B0 with nine stages that outputs 1280-dimensional feature vector.

The NN classifier  consists of two fully connected linear layers with ReLU activation function in between and softmax activation function in the output. The number of hidden neurons was set to 64, while the number output neurons equals 16, which is the number of the categories of tissues to be classified. Initial weights of the NN classifier were randomly sampled from  zero mean normal distribution $N(0, \sigma^2)$ with standard deviation $\sigma=10^{-2}$, while biases were initialized with zeros according to \cite{SimonyanZ14a}. 
\subsubsection{Objective function}
\label{Sec:objective_function}
Training the CNN on the auxiliary task with a cross-entropy loss objective function will force learning image representations that are discriminative for histological structures inherent to the species, organs, and stainings present in the used dataset. We cannot, however, train them to be discriminative for anomalies, since those are not available or even not yet known. One of the approaches for AD is to minimize the volume of hypersphere in the feature space where majority of normal data lives \citep{TaxD04, RuffGDSVBMK18}. Conceptually similar, in \cite{perera2019learning} compactness loss was used to keep low intra-class variance in feature space for normal data. We will also enforce image representations to be compact, while training CNN image representations on the auxiliary task (Fig.~\ref{Fig:1} Top). The compact representations should increase the sensitivity of the system to anomalies. For this purpose we use center-loss $\mathcal{L}_{CL}$ term \citep{WenZL016} in the objective function, along with the multiclass cross-entropy $\mathcal{L}_{CE}$ term
\begin{equation}
\label{Eq:1}
\mathcal{L} = \mathcal{L}_{CE} + \lambda\mathcal{L}_{CL}, 
\end{equation}
where $\lambda$ controls the influence of center-loss. We would like to have  a class of healthy tissues as compact as possible, however, a too large value of $\lambda$ may decrease discriminative properties of learned representations. In our experiments  we set $\lambda=1$.
The multi-class cross entropy is computed as 
\begin{equation}
\mathcal{L}_{CE} = - \sum_{i}^{m} \log P(x_i \in C_{y_i}),
\end{equation}
where $C_{y_i}$ denotes $y_i$-th class, $y_i \in [1,n]$ are true class labels for image representations (feature vectors) $x_i$ within a mini-batch of size $m$. In our case the number of classes $n=16$. The probabilities $P(x \in C_{y_i})$ are computed via the softmax function
\begin{equation}
P(x_i \in C_{y_i}) = \frac{e^{ z_{y_i}(x_i)}}{\sum_{j=1}^{n} e^{z_j(x_i)}}
\end{equation}
with $z_{j}$ being $j$-th entry of the $n$-dimensional output of the last fully connected layer of the NN classifier. 

We define the center-loss as follows
\begin{equation}
\label{Eq:4}
\mathcal{L}_{CL} = \sum_{k \in K \subseteq[1, n]} \frac{\frac{1}{2} \sum_{i=1}^{m}\|x_i-a_k\|^2 \delta(y_i-k)}{\sum_{i=1}^{m}\delta(y_i - k)},
\end{equation} 
where the delta function $\delta$ equals $0$ everywhere except for an argument of zero, where it equals $1$, and $a_k$ is an average feature vector for class $C_k$ within a mini-batch
\begin{equation}
\label{Eq:5}
a_k = \frac{\sum_{i=1}^m x_i \delta(y_i-k)}{\sum_{i=1}^{m} \delta(y_i-k)}
\end{equation}
\textcolor{marked}{In contrast to \cite{WenZL016, perera2019learning}, with the center-loss in Eq.~\ref{Eq:4} we can enforce the compactness of arbitrary classes depending on the subset $K$ over which the summation runs. Additionally, we made the influence of each class independent of its size by proper normalization.}

We optimize Eq.~\ref{Eq:1} by means Stochastic Gradient Descent (SGD) using backpropagation that updates weights $W$ of the CNN and NN classifier at iteration $t$ (backward pass) according to
\begin{equation}
\label{Eq:6}
W^{t+1} = W_t -\mu\frac{\partial \mathcal{L}^t}{\partial W^t},
\end{equation}
where $\mu$ is a learning rate.
We do not directly use Eq.~\ref{Eq:5} to avoid perturbations caused by mislabeled samples as was suggested in \cite{WenZL016}. Instead, at each forward pass of iteration $t$ we update class centers according to
\begin{equation}
\label{Eq:7}
a_k^{t+1} = (1-\beta)a_k^t + \beta \hat{a}_k^{t+1},
\end{equation}
where $\hat{a}_k^{t+1}$ are class new centers computed with Eq.~ \ref{Eq:5} after all weights were updated with Eq.~\ref{Eq:6}\footnote{Eq. 4 with $1$ neglected in denominator and step 6 of an Algorithm 1 in \cite{WenZL016} boil down to Eq.~\ref{Eq:7}.}. $a_k^0$ are initialized with Eq.~\ref{Eq:5} computed with $x_i(W)$ for initial wights $W$. In our experiments in Sec.~ \Ref{Sec:Experiments} we used $\beta=0.5$, even though we have seen only negligible differences in results when we have used $\beta=1$. 

\subsection{Class mix-up augmentation}
\label{Sec:mixup_augmentation}
Histological images carry either textural or color information. Though textural information describes tissue structures that characterize histopathological features, color information is influenced by acquisition environment, concentration of staining reagents that depend on a particular laboratory where histological slides were prepared and digital images were acquired. Moreover, color distribution differences can be striking even between different batches of data prepared in the same laboratory at different times. There is therefore a danger that the classifier learns color patterns common for a particular batch of data instead of diagnostically relevant stained tissue structures. Normalization and augmentation approaches were developed to overcome this problem \citep{TellezLBBBCL19}. The first standardizes the distribution of pixels values during the inference time, while the second increase the variability of pixel  distribution during the training time. For example, hue and/or saturation of pixel values can be randomly shifted. Usually aggressive augmentation is required to improve robustness of the trained classifier to variability between batches of image data \citep{StackeEUL21}. However, since such augmentation is not based on real data, it is hard to tune to gain robustness and simultaneously keep a high performance of a classifier \citep{TellezLBBBCL19}. 

Here we propose a different type of simple augmentation that does not need tuning. Our auxiliary task is a multi-class classification, we therefore can transfer color patterns between different classes (Fig.~\ref{Fig:1}). In this way we increase the  emphasis of a trained classifier on morphology of tissue structures and decrease importance of color that is dependent on laboratory settings at the time of tissue preparation and image acquisition. We do this with the histogram matching technique \citep{GonzalezW18} applied for each of the three color channels separately. 

We do not match histograms between images, but rather find a mapping that matches histograms between sets of images belonging to classes in the training data.   
Prior to training of multi-class classifier, we build the histogram $H_k$ for each class $k \in [1, n]$ in  the training dataset with $n=16$ classes. For pairs of source $k_1$ and target $k_2$ classes we then find a mapping $z=M_{k_1 \rightarrow k_2}(x)$ that maps pixel intensity values $x$ to $z$, such that histogram of mapped pixel intensities of class $k_1$ matches the histogram of class $k_2$. 
Such mappings can be found with the use of Cumulative Distribution Functions (CDF), and in discrete case is computed as $CDF(x)=\sum_{y \leq x} H(y)$.
Since we want to equalize CDFs of source and target classes 
\begin{equation}
\label{Eq:8}
CDF_{k1}(x) = CDF_{k2}(z),
\end{equation}
the desired  mapping becomes
\begin{equation} 
\label{Eq:9}
z = CDF_{k2}^{-1}(CDF_{k1}(x)) = M_{k_1 \rightarrow k_2}(x).
\end{equation}
In practice, since we have only a finite discrete set of intensity values 
we build a look-up table that according to Eq.~\ref{Eq:8} gives closest $z$ for every $x$ from the source class. We use histograms and corresponding CDFs discretized to 256 values. 

\begin{figure}[!htb]
\centering
\includegraphics[scale=.3]{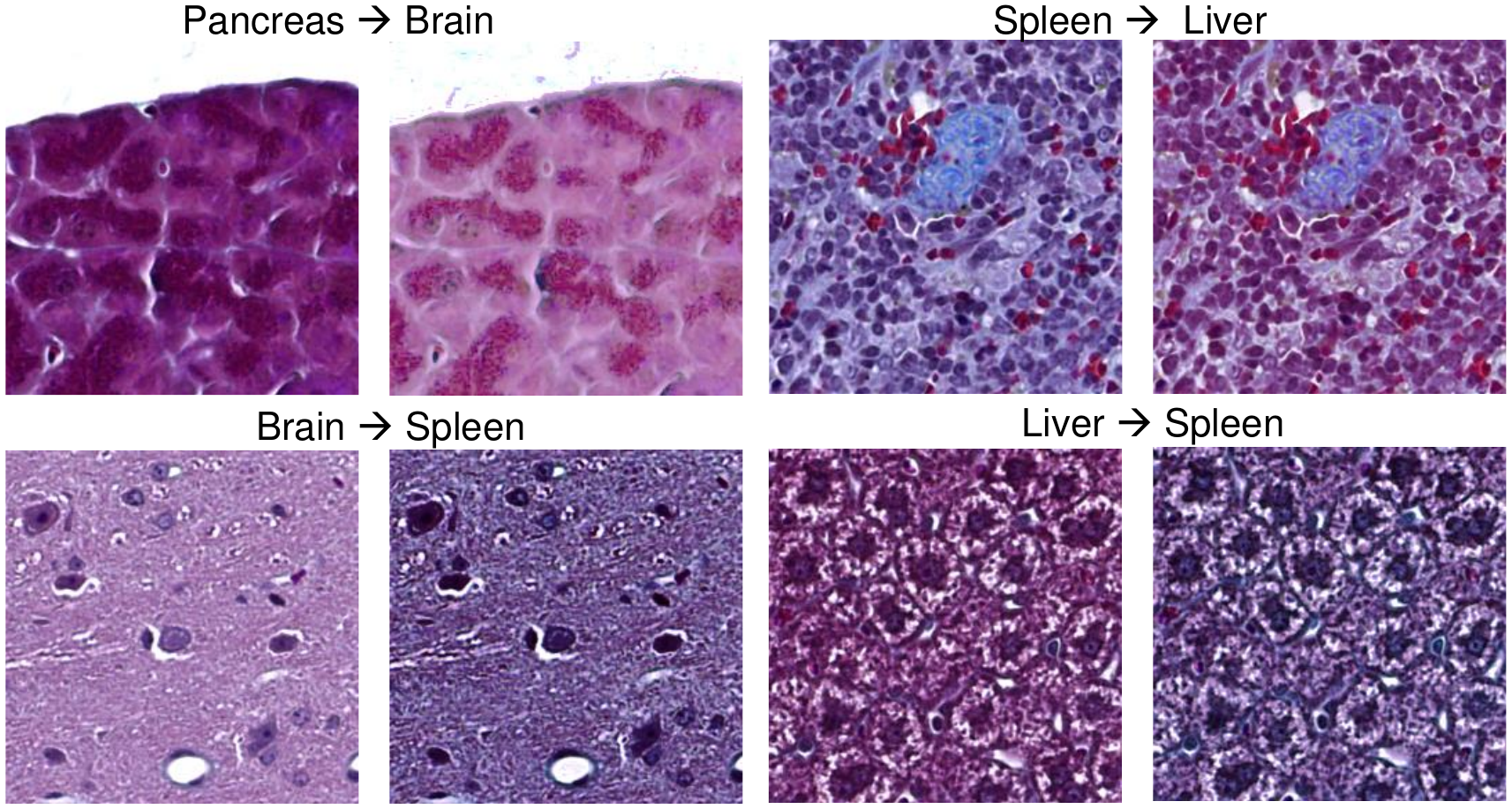}
\caption{\label{Fig:3} Examples of Masson's Trichrome stained tissue images transformed to match color patterns of different tissue classes.}
\end{figure}

During the training time, given an image that belongs to class $k_1$, we randomly choose a destination class $k_2$ and transform pixel intensities of the image with appropriate mapping $M_{k_1 \rightarrow k_2}$. The class $k_2$ is uniformly sampled from $[1, n]$, which also gives a mapping from a particular class to itself ($k_2=k_1$). 
It is important to note that we avoid mapping between classes of different staining (H\&E and Masson's Trichrome), since our classifier is trained to be sensitive to the staining type. \textcolor{marked}{For example, color mix-up augmentation can randomly transfer color shades of mouse pancreas tissue samples stained by MT to color shades of mouse brain samples stained by MT, or transfer color shades of mouse kidney tissue stained by H\&E to color shades of rat liver samples stained with H\&E. Fig.~\Ref{Fig:3} visually exemplifies color pattern transformations for images of MT stained tissue from different (mouse) organs.}

\subsection{One-class classification}

For one-class classification we have used the standard one-class SVM \citep{ScholkopfPSSW01} with the Radial Basis Function (RBF) kernel and margin error $\nu=0.1$, which bounds the fraction of outliers that lie on the wrong side of an SVM margin. Feature vectors that do not lie within the support region of normal healthy data generate anomaly scores with negative values. Once the classifier is trained on image representations of normal tissue of a particular class (e.g. Masson's Trichrome stained mouse liver), it will serve as a detector of abnormal patches for that class, which will be detected by means of simple thresholding of output scores. In Sec.~\Ref{Sec:Experiments} we will evaluate our AD system on finding pathologies in mouse liver tissue.

\section{Experiments}
\label{Sec:Experiments}

In this section we convey a sequence of experiments to evaluate the performance of our approach (\ref{Sec:evaluationNAFLD}), show the importance of learning image representations (\ref{Sec:performance_pretrained}), compare our method to the state-of-the-art anomaly detection approaches (\ref{Sec:SOTAcomparison}), and show the importance of the introduced techniques to the resulted performance (\ref{Sec:ablation_study}).
In Sec.~\ref{Sec:comparison_with_NAS} we study the behavior of our AD method when assessing  Non-Alcoholic
Fatty Liver Disease (NAFLD) as example of a tissue anomaly and compare it with the methods specifically tailored to NAFLD. We also provide a study case that outlines the ability of our approach to detect abnormal alterations in tissue caused by adverse effects of candidate drugs. 

We train and evaluate our AD method using $256 \times 256$ pixel tiles extracted from Whole Slide Images (WSI). WSIs were acquired with a Zeiss AxioScan scanner (Carl 
Zeiss, Jena, Germany) with a $20 \times$ objective at a resolution 0.221 \textmu m/{pixel} from mouse and rat tissue samples\footnote{Only historic data was used. No new animal experiments were done. For historic studies, animals were maintained in accordance with German national guidelines, legal regulations and the guidelines 
of the Association for Accreditation of Laboratory Animal Care. Experiments were performed after permission 
from the Regierungspräsidium Tübingen, Germany.} stained with either H\&E or Masson Trichrome stains according to established protocols.
The WSI were then subsampled with a factor of 1:2, which resulted in 0.442 \textmu m/{pixel} resolution.

To test and compare the performance of AD methods we used a dataset with samples of NAFLD, see details in Sec.~ \ref{Sec:NAFLD_dataset} that will serve as a particular example of anomalous tissue developments.  
 
We implemented all the experiments in the PyTorch deep learning framework \citep{PaszkeGMLBCKLGA19}. Trained models and code are available online\footnote{\href{https://github.com/Boehringer-Ingelheim/anomaly-detection-in-histology}{Boehringer-Ingelheim-anomaly-detection-in-histology}}.  

\subsection{Performance measures}

To evaluate the detection performance of our system we will prefer to use the balanced accuracy measure.
\textit{Balanced accuracy} is an average of detection sensitivity for positives (true positive rate or sensitivity) and for negatives (true negatives rate or specificity). This measure is not sensitive to a proportion of positives and negatives in the test dataset. It is therefore preferable measure when data might be imbalanced, such as in the case of anomaly detection with a small amount of anomalous and the large number of normal samples\footnote{For evaluation of our method in the Sec.~\Ref{Sec:evaluationNAFLD} we have collected a comparable number of anomalies and normal samples.}. We will prefer balanced accuracy over other commonly used measures, such as AUROC and $F_1$ score \citep{SantafeIL15}. One of disadvantages of AUROC is that it summarizes detector's performance over all, also inappropriate, score thresholds, while the disadvantage of  $F_1$ is that it does not care about correct classification of negatives and sensitive to imbalance between negative and positive samples in the tests set \citep{BrabecKFM20}.
\textcolor{marked}{However, in Sec.~\ref{Sec:SOTAcomparison}, we will use AUROC and $F_1$ score to compare our approach with DPA \citep{ShvetsovaBFSD21} and six other SOTA AD methods from Anomalib benchmark \citep{AkcayAB18}, where results are reported with AUROC and $F_1$ measures. }

\subsection{Training image representations}
\label{Sec:dataset_auxiliary}
To train image representations using an auxiliary task as described in Sec.~\ref{Sec:Auxiliary} (See Top of Fig.~\ref{Fig:1}) we collected a large dataset of histological $256 \times 256$, 0.442  \textmu m/{pix.} tiles extracted from Whole Slide Images (WSI) of healthy animals tissue. Seven organs (liver, brain, kidney, heart, lung, pancreas, spleen), two animal species (rat and mouse), and two types of staining (Masson Trichrome and H\&E) comprises the collected dataset. Overall we have 16 categories. Note that not all combinations of specie, organ, and staining was available for us (rat
tissue was available only for liver). Around 7000 tiles from each of the 16 categories is sampled from the dataset. During the training of FC-NN and CNN models and we will use 6 different random seeds that randomizes the choice of the 7000 sampled tiles and the order of their feeding to the network. During the test time we will provide mean and standard error values over 6 trained models, which will allow us more reliable performance comparisons. 
Note that the extracted from WSI tiles were not curated by a human and may occasionally contain various artifacts or wrong tissue. We made this dataset of healthy tissue publicly available \citep{dataset} such that training representations of histological images for our AD approach can be reproduced.

We train FC-NN and CNN models over 15 epochs using stochastic gradient descent optimization with learning rate 1e-3, momentum 0.9, and with batches of size 64. We choose the best model obtained during the training based on a validation set (around 700 images per class) withheld from the dataset. 

The CNN encoder was initialized with weights obtained from pre-training on ImageNet, while fully connected layers of the classifier were randomly sampled from  zero mean normal distribution $N(0, \sigma^2)$ with standard deviation $\sigma=10^{-2}$ and biases were initialized with zeros. 

We used the categorical cross-entropy objective function with a center-loss term (see Sec.~\ref{Sec:objective_function}) that forces compactness of a single class that is a tissue class of interest  for AD system (either H\&E or Masson Trichrome stained mouse liver tissue, \textcolor{marked}{sometimes also referred to as target class}). This implies that the subset $K$ in Eq.~ \ref{Eq:4} is equal to a single number corresponding to the class of interest.

All the images were normalized to have a zero mean over the training dataset for each color channel. No normalization of standard deviation of image values was done.

After class-mixup augmentation (see Sec.~ \ref{Sec:mixup_augmentation} for details), brightness and contrast augmentations were additionally applied\footnote{\label{FN:ColorJitter}We have used Torchvision extension of PyTorch for this purpose, namely \href{https://pytorch.org/vision/stable/generated/torchvision.transforms.ColorJitter.html}{ColorJitter.}}. Brightness and contrast were randomly altered with a factor uniformly chosen in the range [0.8, 1.2].   

\subsection{Training one-class classifier}
For one-class classification we have used the standard one-class SVM \citep{ScholkopfPSSW01} with the Radial Basis Function (RBF) kernel and a margin error $\nu=0.1$. During training one-class classifier (See Middle of Fig.~\ref{Fig:1}) we perform brightness, contrast, saturation and mild hue augmentations\footnotemark[5]
with factors uniformly sampled from [0.8, 1.2], [0.8, 1.2], [0.4, 1.6], [-0.05, 0.05], respectively. All the images were normalized to have a zero mean over the training dataset for each color channel. No normalization of standard deviation of image values was done. We carried experiments and reported results for two types of staining Masson Trichrome and H\&E. For each type of staining one-class classifier was trained on the corresponding dataset of liver mouse Masson Trichrome or H\&E stained tissue, which were also used for training image representations (see Sec.~\ref{Sec:dataset_auxiliary})

\subsection{The dataset for performance evaluation}
\label{Sec:NAFLD_dataset}
To evaluate the performance of our approach for the detection of abnormal tissues, we have collected $256 \times 256$, 0.442  \textmu m/{pix.} tiles that were taken from histological WSIs of mouse liver tissue stained with Masson Tichrome and H\&E. We have created three datasets, which we made publicly available \citep{dataset}. The first is the training dataset with samples of healthy mouse liver tissue, which is used to train one-class classifier. It consist of approximately 7000 tiles for each type of staining. The tiles were extracted from 204 WSIs for Masson Trichrome staining and 347 WSIs from H\&E staining.  The second is a test dataset of healthy mouse liver tissue samples that consist of around 2200 tiles with H\&E stained tissue and about 2400 tiles with Masson Trichrome stained tissue. The tiles were extracted from 17 WSIs for each of the two stainings. The third is a test dataset of anomalous tissue from mice with NAFLD condition. The tissue can have a few different anomalous tissue alterations, namely inflammation, ballooning, steatosis, and fibrosis. It contains about 2200 tiles with H\&E stained tissue and about 2400 tiles with Masson Trichrome stained tissue. The tiles were extracted from 58 WSIs for Masson Trichrome staining and 57 WSIs fro H\&E staining. The majority of the slides with tissue stained with one method (Masson Trichrome or H\&E) have a matching slide with close tissue cut (3-4 \textmu m distance) stained with the other method. The labels (anomalous NAFLD or healthy) for all tiles were carefully verified by an experienced pathologist. The test datasets have no common WSIs neither from the one-class classifier training dataset described above nor from the dataset for training image representations on the auxiliary task (see Sec.~\ref{Sec:dataset_auxiliary}).

\subsection{Performance evaluation based on NAFLD anomalies}
\renewcommand{\arraystretch}{1.3}
\label{Sec:evaluationNAFLD}
We evaluate the performance of our AD system (see Fig.~\ref{Fig:1}C) using encoders based on a few different backbone CNN architectures. Here we report the performance using the balanced accuracy  and provide the mean over the results obtained from six trained CNN encoders that were trained with six seeds (0, 100, 200, 300, 400, 500). We also provide a standard error for the estimated means. The results are summarized in Table~\ref{Tab:1} for both Masson Trichrome and H\&E staining. The average between these performances is reported in the last row. 

DenseNet-121, EfficientNet-B0, and EfficientNet-B2 backbone CNN are reported within two versions, the full CNN and truncated CNN with smaller size of output feature vectors, which may improve generalization of the trained network to unseen data and reduce overfitting to the used training dataset. Additionally, it makes training faster. The truncated version of EfficientNet-B0 and EfficientNet-B2 is obtained by removing the last $9^{th}$ stage with $1\times1$ convolutional layer, the truncated version of DenseNet-121 is obtained by removing the fourth (last) dense block, 
such that the number of the output channels is reduced to the number shown in corresponding entry of the row refereed to as Feature Vector (FV) size. 

From Table~\ref{Tab:1} one can see that the best (average) performance of AD system, $95.86\%$, was obtained with truncated EfficientNet-B0 with 320 output features. The truncated version of EfficientNet-B0 as well as truncated versions of all three aforementioned networks performed better than their full versions. This points to possible overfitting the auxiliary task that mostly  occurs due to learned weights in last layers of the CNN encoder, which reduces performance of the AD system. Note that Table~\ref{Tab:1} does not report the performance of the supervised classifier for the auxiliary task (Fig.~\ref{Fig:1} A), which typically reaches accuracy above $97\%$, but the performance of the AD system (Fig.~\ref{Fig:1} C). 

\begin{figure}[!htb]
\centering
\includegraphics[scale=0.75]{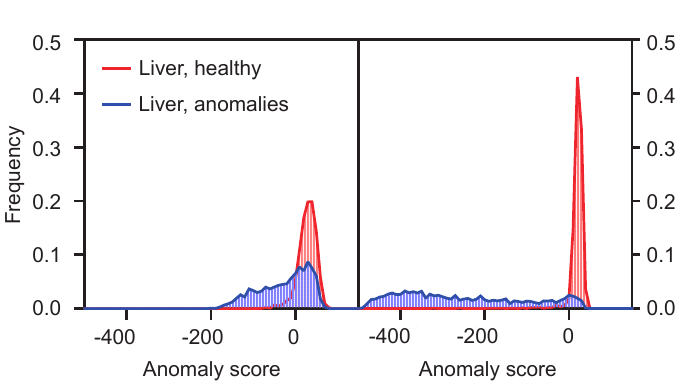}
\caption{\label{Fig:anomalyscores} Distribution of anomaly scores generated by the AD system (output of one-class classifier) for  the test dataset of tissue stained with H\&E. The AD system uses 320 dimensional image representations generated by \textbf{Left:} pre-trained (on ImageNet) EfficientNet-B0 and by \textbf{Right:} EfficientNet-B0 trained with the proposed techniques on our auxiliary task (BIHN model).}
\end{figure}

\begin{figure}[!htb]
\centering
\includegraphics[scale=1.3]{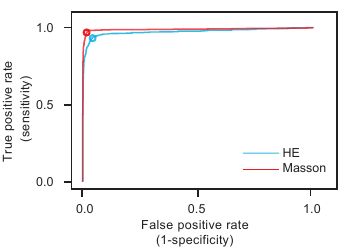}
\caption{\label{Fig:ROC} \textcolor{marked}{ROC curve for detection of tiles with NAFLD using BIHN models trained for Masson Trichrome or H\&E tissue staining. A circle on the curves correspond to the working point of the one-class classifier.}}
\end{figure}

The distribution of anomaly scores of the AD system (the output of one-class SVM classifier) equipped with truncated EfficientNet-B0 that was trained on our auxiliary task using the proposed procedures is shown in Fig.~\ref{Fig:anomalyscores} on the right. The corresponding Receiver Operating Characteristic (ROC) curve of the AD system is shown in Fig.~\ref{Fig:ROC}. For comparison, the distribution of anomaly scores obtained from the AD system that uses the same neural network architecture but trained on ImageNet is shown in the Fig.~\ref{Fig:anomalyscores} on the left. For the chosen working point of our AD system (see the circle on Fig.~\ref{Fig:ROC}) we obtain sensitivity about 0.98, specificity about 0.97 for the case of Masson's Trichrome staining and sensitivity about 0.93, specificity about 0.96 for the case of H\&E staining.

In addition to CNN architectures shown in  Table~\ref{Tab:1} we have experimented with visual transformer ViT-B-32 \citep{DosovitskiyB0WZ21} that outputs 768-dimensional feature vectors\footnote{In contrast to the tested CNN architectures with $256 \times 256$ input images, for visual transformer input images are $224 \times 224$ pixels.}. However, it performed worse than the majority of CNN architectures giving an average performance $93.44 \pm 0.44$.

In our further experiments we will be using our best performing architecture EfficientNet-B0-320 and will call trained on the auxiliary task CNNs with this architecture Boehringer Ingelheim Histological Network (BIHN) models.

\begin{table*}[!htb]
\caption{\label{Tab:1}Performance of the proposed AD method for different CNN backbones, measured with Balanced Accuracy in $\%$; \textcolor{marked}{standard errors are given.}}
\scriptsize
\centering
\begin{tabularx}{\textwidth}{XXXXXXXXXXXX}
\hline
  & VGG-11 & VGG-16 & VGG-19 & ResNet-18  & DenseNet-121 & DenseNet-121 & EfficientNet-B0 & EfficientNet-B0(BIHN$^@$) & EfficientNet-B2 & EfficientNet-B2 & ConNeXt-T \\
\hline
FV$^+$ size & $512$ & $512$ & $512$ & $512$ & $512$ & $1024$ & $1280$ & $320$ & $1480$ & $352$& $768$\\
Learned param. & 9,254,352 & 14,748,560 & 20,058,256 & 11,723,384 & 4,828,624 & 7,020,496 & 4,090,572 & 3,617,612 & 7,792,210 & 7,226,898 & 27,868,848 \\
\hline
\hline
MT$^*$ &     \mbox{$94.60 \pm 0.24$} & \mbox{$92.08 \pm 0.76$} & \mbox{$90.03 \pm 0.95$} & \mbox{$93.89 \pm 0.80$} & \mbox{$96.12 \pm 0.41$} & \mbox{$96.82 \pm 0.22$} & \mbox{$97.17 \pm 0.26$} & \mbox{$\textbf{97.51} \pm 0.12$} & \mbox{$96.36 \pm 0.25$} & \mbox{$95.82 \pm 0.20$} & \mbox{$96.52 \pm 0.19$}\\
\hline
H\&E$^\#$  & \mbox{$93.59 \pm 0.54$} & \mbox{$92.67 \pm 1.46$} & \mbox{$91.83 \pm 0.93$} & \mbox{$\textbf{95.35} \pm 0.29$}  & \mbox{$94.37 \pm 0.21$} & \mbox{$93.96 \pm 0.39$} & \mbox{$93.67 \pm 0.26$} & \mbox{$94.20 \pm 0.12$} & \mbox{$93.45 \pm 0.12$} & \mbox{$94.95 \pm 0.23$} & \mbox{$93.81 \pm 1.05$}\\           
\hline
Average    & \mbox{$94.10 \pm 0.39$} & \mbox{$92.37 \pm 1.12$} & \mbox{$90.93 \pm 0.94$} & \mbox{$94.62 \pm 0.54$}  & \mbox{$95.25 \pm 0.31$} & \mbox{$95.37 \pm 0.30$} & \mbox{$95.42 \pm 0.26$} & \mbox{$\textbf{95.86} \pm 0.12$} & \mbox{$94.91 \pm 0.19$} & \mbox{$95.39 \pm 0.22$} & \mbox{$95.17 \pm 0.62$}\\     

\hline

\multicolumn{12}{@{}l}{$^*$MT -- Masson's Trichrome staining \qquad 
$^{\#}$ H\&E -- Hematoxylin and Eosin staining \quad $^+$ FV -- Feature Vector \quad $^@$ BIHN -- Boehringer Ingelheim Histological Network}

\end{tabularx}
\end{table*}

\subsubsection{Advantage of learning representations for histological images}
\label{Sec:performance_pretrained}
Here we provide a comparison of our AD system, which is composed of a CNN encoder and a one-class classifier, with the system of a similar architecture but with the CNN encoder trained on ImageNet without further adaptation to histopathological data using our auxiliary task. Table~\ref{Tab:2} summarizes the performances of AD for a set of CNN architectures (the same set as in Sec.~ \ref{Sec:evaluationNAFLD}) trained on the ImageNet dataset. Put attention that in this table we report balanced accuracy measures with no standard error as in Sec.~\ref{Sec:evaluationNAFLD}, where we also reported the standard error of the estimated mean over a few CNN encoders trained on random subsets of histological images.

Comparing Tables~\ref{Tab:1} and  \ref{Tab:2} one can see an essential increase in the performance when CNN encoders were trained on histological data. For example, for the best performing architecture EfficientNet-B0-320 the balanced accuracy increases in $26\%$. The results emphasize the importance of tuning the neural networks to a particular target domain. Even though recently it was shown that using transfer learning concept for AD \citep{BergmannFSS20, RippelMM20}, where ImageNet trained neural networks encode images with feature vectors representations, can achieve state of the art results. However, for many cases where target domain diverges from natural images, like in our case of histopathological images, tuning neural networks to the target domain is critical.

\begin{table*}[!htb]
\caption{\label{Tab:2}One-class SVM AD performance when ImageNet trained CNNs were used as an image encoder, measured with Balanced Accuracy in $\%$}
\scriptsize
\centering
\begin{tabularx}{\textwidth}{XXXXXXXXXXXX}
\hline
  & VGG-11 & VGG-16 & VGG-19 & ResNet-18  & DenseNet-121 & DenseNet-121 & EfficientNet-B0 & EfficientNet-B0 & EfficientNet-B2 & EfficientNet-B2 & ConNeXt-T \\
\hline
FV$^+$ size & $512$ & $512$ & $512$ & $512$ & $512$ & $1024$ & $1280$ & $320$ & $1480$ & $352$& $768$\\
\hline
\hline
MT$^*$ & $71.64$ & $68.99$ & $69.46$ & $70.68$ & $71.64$ & $76.10$ & $73.99$ & $\textbf{77.21}$ & $73.65$ & $73.75$ & $64.89$\\
 
\hline
H\&E$^\#$  & $73.91$ & $72.91$ & $73.96$ & $72.02$  & $73.42$ & $\textbf{75.76}$ & $72.17$ & $74.90$ & $72.28$ & $71.78$ & $65.51$\\ 
\hline
Average  & $72.77$ & $70.95$ & $71.71$ & $71.35$  & $72.53$ & $75.93$ & $73.08$ & $\textbf{76.05}$ & $72.97$ & $72.76$ & $65.20$\\

\hline

\multicolumn{12}{@{}l}{$^*$MT -- Masson's Trichrome staining \qquad 
$^{\#}$ H\&E -- Hematoxylin and Eosin staining \quad $^+$ FV -- Feature Vector}

\end{tabularx}
\end{table*}

\subsubsection{Comparison with AD methods}
\label{Sec:SOTAcomparison}

We compare the developed AD method with recent methods for anomaly detection with implementations available from the Anomalib \citep{abs-2202-08341web} benchmark \textcolor{marked}{and with the official implementation of Deep Perceptual Autoencoder (DPA) \citep{ShvetsovaBFSD21}}. We also compare the developed AD method with the classical one-class SVM \citep{ScholkopfPSSW01} that uses feature representations obtained with  a CNN pre-trained on ImageNet \citep{DengDSLL009}.

\textcolor{marked}{All the methods were trained on healthy mouse liver tissue training dataset and tested on healthy tissue (normals) and tissue with NAFLD (anomalies) test datasets. The datasets are described in Sec.~\ref{Sec:NAFLD_dataset}}. The results are summarized in Table~\ref{Tab:3}. The table shows detection performance of the approaches measured with the Area Under ROC curve (AUROC) and the $F_1$ score\footnote{Best $F_1$ score over all thresholds is calculated for the methods from Anomalib and for DPA, while for our method and for one-class SVM with ImageNet pre-trained CNN,  the standard $F_1$ score is calculated on thresholded output of one-class classifier. } on the NAFLD dataset (see Sec.~\ref{Sec:NAFLD_dataset}). \textcolor{marked}{Standard errors are given along with estimated performances.} The table summarizes the experiments with H\&E and Masson trichrome stained tissues and shows an average performance between the experiments with two stains. 

Columns 1-6 in Table~\ref{Tab:3} show performance of anomaly detection algorithms from the Anomalib benchmark.
For the methods implemented in Anomalib we used hyperparameters reported in the corresponding paper (and usually set as default parameters in the configuration files of Anomalib benchmark). Below we explicitly mention only a few important configuration parameters when they are different from the optimal configurations reported in the corresponding papers or when a few configuration options were reported.
In the PatchCore algorithm \citep{roth2022towards}  we used $10\%$ subsampling rate and features from the third block of WideResNet50 backbone. We were not able to use features from three blocks, which gave a bit better performance in the original paper, because the PatchCore is memory intensive and the Anomalib implementation encounter an run-time issue in this case.
In the PaDiM algorithm \citep{defard2021padim} we used features from first three blocks of the ResNet18 backbone. For DFM \citep{abs-1909-11786} we used features from the third block of ResNet50 backbone and feature dimensionality was reduced keeping $99.5\%$ of the original variance. 
For CFLOW \cite{GudovskiyIK22} we used features from second and third blocks of WideResNet50. We did not use features from three blocks, because the training becomes extremely slow.
For STFPM \citep{WangHD021} we used features from the first to third blocks\footnote{The authors of the corresponding paper count first simple convolutional layer as the first block, therefore it is second to fourth blocks as reported in the corresponding paper.}.

\textcolor{marked}{Column 7 shows the performance of the DPA method \citep{ShvetsovaBFSD21} that was trained on full resolution $256 \times 256$ images. The reported results correspond to the weakly supervised setting of the method, where hyperparameters are optimized using examples of both anomalous and normal samples (from our test dataset). Particularly, the reduction of the bottleneck of the auto-encoder from the default 64 to 16 channels (latent dimensions in the DPA configuration files) improved the performance measures. Since, it is unfair to the other methods, we also report here the performance for default parameters (unsupervised setting of the method, $256 \times 256$ images), which is AUROC = $93.11 \pm 0.35$, F$_1$ = $86.32 \pm 0.39$ for average performance between MT and H\&E models and datasets. It is interesting to note that the reconstruction quality of the auto-encoder essentially deteriorated for the optimized (reduced) number of bottleneck channels, which means that  good image reconstruction should not necessary imply high performance of anomaly detection. The progressive growing mode of training DPA did not improve the results (it, however, speeds up training), as well as normalization of the images using the \cite{7460968} method.}


We trained image representations on an auxiliary task as proposed in our paper using the EffcientNet-B0-320 CNN architecture (BIHN models). 
The  results for AD with BIHN models are reported in $9^{th}$ (last) column\footnote{\textcolor{marked}{Performance of our anomaly detector is also reported in the $8^{th}$ column in Table~\ref{Tab:1} but using the balanced accuracy measure.}}. The same architecture, EffcientNet-B0-320, trained on ImageNet and used together with one-class SVM gives results reported in $8^{th}$ column.  

\textcolor{marked}{The mean and standard error were calculated based on 6 trials using six different seeds. 
For our method we correspondingly trained 6 different BIHN models. Put attention that in $8^{th}$ column of Table~\ref{Tab:3} we do not report standard error values, as in all other columns, because we used the single pre-trained EfficientNet-B0-320 network.}

As Table~\ref{Tab:3} shows, our approach outperforms, with a confidence, all the other tested approaches for detection of NAFLD. Our BIHN model followed by a standard one-class SVM (see Fig.~\ref{Fig:1}) achieves $3.5\%$ and $8.1\%$ higher performance than the closest method, DPM \citep{ShvetsovaBFSD21}, measured with AUROC and $F_1$ measures, respectively.
One can also see that the same architecture, EficientNet-B0-320 CNN encoder followed by one-class SVM, when trained on ImageNet, achieves substantially lower performance, which emphasizes critical value of learning image representations.

\begin{table*}[!htb]
\caption{\label{Tab:3}Comparison of anomaly detection methods.}
\scriptsize
\centering
\begin{tabularx}{\textwidth}{XXXXXXXXXXX}
\hline
& & GANomaly & CFLOW & PaDiM & STFPM  & PatchCore & DFM & \textcolor{marked}{DPA, weak supervision} & OC-SVM, Eff.Net feat. & BIHN \\
 & &  {\scriptsize \cite{AkcayAB18}} &  {\scriptsize \cite{GudovskiyIK22}} & {\scriptsize \cite{defard2021padim}} &  {\scriptsize\cite{WangHD021}} &  {\scriptsize \cite{roth2022towards}}&  {\scriptsize \cite{abs-1909-11786}} & {\textcolor{marked}{\scriptsize \cite{ShvetsovaBFSD21}}} & {\scriptsize\cite{ScholkopfPSSW01}} & {\scriptsize ours}\\

\hline
\hline
MT$^*$ &     AUROC $\%$ & $84.26 \pm 2.63$ & $75.36 \pm 1.24$ & $79.63 \pm 0.45$ & $80.99 \pm 0.70$  & $85.46 \pm 0.02$ & $92.28 \pm 0.18$  & $94.26 \pm 0.29$ & $85.53$ & $\textbf{99.03} \pm 0.12$ \\
     & F$_1$ score $\%$ & $77.53 \pm 3.27$ & $71.74 \pm 0.53$ & $73.81 \pm 0.34$ & $77.92 \pm 0.38$  & $79.50 \pm 0.03$ & $85.02 \pm 0.17$ & $87.42 \pm 0.38$ & $72.13$ & $\textbf{97.51} \pm 0.12$ \\     
\hline
H\&E$^\#$  & AUROC $\%$ & $59.59 \pm 2.03$ & $74.29 \pm3.51 $ & $86.33 \pm 0.38$ & $83.24 \pm 0.88$  & $86.93 \pm 0.03$ & $92.69 \pm 0.0$ & $95.06 \pm 0.21$ & $78.63$  & $\textbf{97.33} \pm 0.14$ \\
     & F$_1$ score  $\%$ & $66.45 \pm 0.0$  & $71.80 \pm 1.1$  & $78.30 \pm 0.30$ & $80.03 \pm 0.50$  & $79.85 \pm 0.06$ & $85.40 \pm 0.0$ & $88.07 \pm 0.30$ & $69.61$ & $\textbf{94.09} \pm 0.13$ \\     
\hline
Average    & AUROC $\%$ & $71.92 \pm 2.33$ & $74.83 \pm 2.37$ & $82.98 \pm 0.42$ & $82.12 \pm 0.79$  & $86.20 \pm 0.03$ & $92.49 \pm 0.09$ & $94.66 \pm 0.25$ & $82.08$  & $\textbf{98.18} \pm 0.10$ \\
    & F$_1$ score $\%$ & $71.99 \pm 1.64$ & $71.77 \pm 0.82$ & $76.05 \pm 0.32$ & $78.97 \pm 0.44$  & $79.67 \pm 0.04$ & $85.21 \pm 0.09$ & $87.75 \pm 0.34$ & $70.87$  & $\textbf{95.80} \pm 0.13$ \\
\hline

\multicolumn{10}{@{}l}{$^*$MT -- Masson's Trichrome staining \qquad 
$^{\#}$ H\&E -- Hematoxylin and Eosin staining}

\end{tabularx}
\end{table*}

\subsubsection{Comparison with self supervised learning approaches}
\label{sec:SSL}
Self-supervised learning (SSL) was shown to create powerful feature representations for various computer various tasks \citep{8237488,ericsson2022self}. In our work, to learn suitable feature representations we used fully supervised training\footnote{Our approach for learning feature representations can also be coined self-supervised because for self-supervision we use an auxiliary information without the need to manually collect labels \citep{8237488,kobayashi2022self}.} on the proposed auxiliary task. In Table~\ref{Tab:SSL} we compare both approaches, our auxiliary task versus recent contrastive, SimCLR \citep{chen2020simclr}, and non-contrastive, SimSiam \citep{chen2021exploring}, SSL methods to learn feature representations for the anomaly detection in histopathological images\footnote{Best $F_1$ score over all thresholds is calculated for the SSL methods, while for our method the standard $F_1$ score is calculated on thresholded output of one-class classifier.}. These two methods have shown a good performance in the comparative SSL study in the domain of histopathology \citep{voigt2023investigation}. SSL was applied to the same training data that was used for our auxiliary task, but without the labels (organs, species). Training and testing for H\&E and Masson Trichrome stained tissue was performed separately. 

We used implementations of SimCLR and SimSiam publicly available at \href{https://docs.lightly.ai/self-supervised-learning}{Lightly SSL}, while adopting the optimal parameters from \cite{voigt2023investigation}. Namely, for SimCLR we trained for 80 epochs with batch size 512, initial learning rate 6e-3, and a weight decay 1e-5, using the LAMB optimizer \citep{YouLRHKBSDKH20}. For SimSiam we trained for 50 epochs with batch size 256, an initial learning rate of 0.5, a momentum of 0.5, and weight decay 1e-4, using the stochastic gradient descent (SGD) optimizer. In both cases we have used the cosine
 decay schedule \citep{LoshchilovH17} for learning rate and ImageNet pre-trained ResNet-18 deep neural network. We used the standard augmentations described in the corresponding papers, but added vertical flip due to the nature of histopathological images. Adding rotation augmentation did not change the performance of the learned representations. 

Similarly to \cite{voigt2023investigation}, in our comparison SimSiam outperformed SimCLR. However, its performance is considerably lower than the performance of feature representations learned with our auxiliary task, as can be seen from  Table~\ref{Tab:SSL}. We noticed that applying SSL on target class only (liver mouse tissue instead of all organs and species) reduced performance of SSL.

We also compared feature representations obtained with SSL, ImageNet pre-trained neural network, and supervised learning on our auxiliary task, when all the methods used the neural network of the same ResNet-18 architecture and the anomaly detection performance was measured using balanced accuracy\footnote{For SimCLR and SimSiam we set margin error $\nu$ of one-class SVM to 0.3, which gave us the best performance for the SSL methods.}. From Table~\ref{Tab:SSL_ba} it can be seen that SSL was able to improve performance over the ImageNet pre-trained network, but falls far behind the performance of the features learned from our auxiliary task. The results show that with readily available labels for histopathological data from different organs, species, and stained with different reagents, fully supervised training allows to learn powerful features that are considerably better then when using self-supervised learning techniques. Both contrastive and non-contrastive self-supervision forces feature similarity for images after applied transformations, but not similarity and dissimilarity for images within and between histopathological tissue classes. 

\begin{table}[tb]
\caption{\label{Tab:SSL}Comparison with Self-Supervised Learning (SSL) methods.}
\scriptsize
\centering
\begin{tabularx}{\columnwidth}{XXXXX}
\hline
& & simCLR & simSiam & auxiliary task \\
 & &  {\scriptsize \cite{chen2020simclr}} &  {\scriptsize \cite{chen2021exploring}} & {\scriptsize BIHN model}\\

\hline
\hline
MT$^*$ &     AUROC $\%$ & $81.10 \pm 0.54$ & $85.41 \pm 0.94$ & $\textbf{99.03} \pm 0.12$ \\
     & F$_1$ score $\%$ & $74.11 \pm 0.48$ & $77.81 \pm 1.02$ & $\textbf{97.51} \pm 0.12$ \\     
\hline
H\&E$^\#$  & AUROC $\%$ & $80.36 \pm 0.40$ & $82.86 \pm 0.75$  & $\textbf{97.33} \pm 0.14$ \\
     & F$_1$ score  $\%$ & $74.46 \pm 0.23$ & $74.48 \pm 0.65$ & $\textbf{94.09} \pm 0.13$ \\     
\hline
Average    & AUROC $\%$ & $80.73 \pm 0.47$ & $84.13 \pm 0.84$  & $\textbf{98.18} \pm 0.10$ \\
    & F$_1$ score $\%$ & $74.29 \pm 0.35$ & $76.14 \pm 0.84$  & $\textbf{95.80} \pm 0.13$ \\
\hline

\multicolumn{5}{@{}l}{$^*$MT -- Masson's Trichrome staining  
$^{\#}$ H\&E -- Hematoxylin and Eosin staining}

\end{tabularx}
\end{table}

\begin{table}[tb]
\caption{\label{Tab:SSL_ba} Performance of SSL methods measured with balanced accuracy in $\%$, when all the methods use the ResNet-18 architecture.}
\scriptsize
\centering
\begin{tabularx}{\columnwidth}{XXXXX}
\hline
  &  ImageNet pre-trained & SimCLR & SimSiam & Auxiliary task  \\
\hline

MT$^*$ & $70.68$ & $73.15 \pm 0.16$ & $80.22 \pm 1.71$ & $\textbf{93.89} \pm 0.80$  \\
 
\hline
H\&E$^\#$  & $72.02$ & $73.19 \pm 0.49$ & $77.23 \pm 0.73$ & $\textbf{95.35} \pm 0.29$   \\ 
     
\hline
Average  & $71.35$ & $73.17 \pm 0.33$ & $78.72 \pm 1.22$ & $\textbf{94.6} \pm 0.54$   \\

\hline

\multicolumn{5}{@{}l}{$^*$MT -- Masson's Trichrome staining  
$^{\#}$ H\&E -- Hematoxylin and Eosin staining}

\end{tabularx}
\end{table}

\subsubsection{Ablation Study}
\label{Sec:ablation_study}

Here we will study the importance of the techniques that we proposed to learn feature representations for histological images. Namely, we will study the influence of the class mix-up augmentation (Sec.~\ref{Sec:mixup_augmentation}), the center-loss term in the objective function (Sec.~\ref{Sec:objective_function}), and the complexity of the auxiliary classification task (Sec.~\ref{Sec:Auxiliary}) on the AD performance.  

The ablation study results are summarized in Tab.~\ref{Tab:4}. The $1^{st}$ column shows  the performance of our system \textcolor{marked}{(all the proposed techniques are implemented for training BIHN models).}
Switching off the class mix-up augmentation decreases (average) AD performance in $3.3\%$ ($2^{nd}$ column) or increases the classification error (1-balanced accuracy) in $78\%$. Using instead of the class mix-up the standard way to augment color images by means of random variations in hue and saturation results in even slightly worse than without augmentation performance ($3^{d}$ column).

Excluding the center loss term from objective function results in $3.4\%$  decrease in performance ($4^{th}$ column), or increase in classification error  in $78\%$. We implemented the center-loss that forces compactness of a single class of interest (liver, mouse, particular staining) out of 16 classes in the auxiliary task. We also checked performance of AD when center-loss forces compactness of all the classes in the auxiliary task ($5^{th}$ column). Though, the performance in this case is higher than without center-loss term, it is inferior to the case with center-loss applied to the single class of interest.  

\textcolor{marked}{The $6^{st}$ column shows an essential drop in the performance of AD when the target class (mouse, liver, MT/H\&E) is excluded from the categories of the auxiliary task.  
$7^{st}$ and $8^{th}$ columns show AD performance when the number of categories to be classified in the auxiliary task were reduced to 14 and 7, respectively. Namely, for the case of 14 categories we excluded rat classes, while for the case of 7 categories we excluded classes of tissue with the staining different to the target category.}
These columns show that performance of AD drops inversely proportional to the number of categories, i.e. performance of the AD is proportional to the complexity of the auxiliary task.
This suggests that increasing  the complexity of the auxiliary task may further increase the performance of our AD method.


\begin{table*}[!htb]
\scriptsize
\caption{\label{Tab:4} Performance of the proposed AD when auxiliary task training procedure was altered. The performance is measured with balanced accuracy in $\%$; \textcolor{marked}{standard errors are given.}}
\scriptsize
\centering
\begin{tabularx}{\textwidth}{XXXXXXXXX}
\hline
  & Proposed training procedure & No mix-up & \mbox{Hue-sat.} augmentation, no mix-up  & No center-loss  & Center-loss for all categories & \textcolor{marked}{Target cat. is excluded, center-loss for all cat.} & Number of categories reduced to 14& Number of categories reduced to 7\\
\hline
\hline
MT$^*$ &     \mbox{$\textbf{97.51} \pm 0.12$} & \mbox{$94.39 \pm 0.21$} & \mbox{$88.87 \pm 0.38$} & \mbox{$94.84 \pm 0.21$} & \mbox{$95.32 \pm 0.24$} & \mbox{$84.79 \pm 0.41$} & \mbox{$95.75 \pm 0.14$} & \mbox{$93.26 \pm 0.23$} \\
\hline
H\&E$^\#$  & \mbox{$\textbf{94.20} \pm 0.12$} & \mbox{$90.92 \pm 0.15$} & \mbox{$90.51 \pm 0.26$} & \mbox{$90.40 \pm 0.61$}  & \mbox{$92.04 \pm 0.42$} & \mbox{$83.54 \pm 0.76$} & \mbox{$92.98 \pm 0.13$} & \mbox{$92.69 \pm 0.22$} \\           
\hline
Average    & \mbox{$\textbf{95.86} \pm 0.12$} & \mbox{$92.65 \pm 0.18$} & \mbox{$89.69 \pm 0.32$} & \mbox{$92.62 \pm 0.41$}  & \mbox{$93.68 \pm 0.33$} & \mbox{$84.17 \pm 0.59$} & \mbox{$94.36 \pm 0.14$} & \mbox{$92.97 \pm 0.22$} \\     

\hline

\multicolumn{9}{@{}l}{$^*$MT -- Masson's Trichrome staining \qquad 
$^{\#}$ H\&E -- Hematoxylin and Eosin staining }

\end{tabularx}
\end{table*}

%
%
%

\subsection{Comparison with methods tailored for assessment of NAFLD}
\label{Sec:comparison_with_NAS}

\textcolor{marked}{We analyzed liver tissue samples stained with Masson's Tricrhome (MT) from mice with Non-Alcoholic Fatty Liver Disease (NAFLD) using our BIHN model (correspondingly trained on healthy MT  stained tissue).} Here, the mice were treated with $CCl_4$ (carbon tetrachloride), which leads to the development of liver pathology similar to NAFLD. The livers of the mice were analyzed immediately after $CCl_4$ administration and after a recovery period of 4, 8, and 12 weeks.

Two methods tailored for the quantification of the NAFLD pathology are compared with our anomaly detector. First, the immunohistochemical staining of alpha smooth muscle actin ($\alpha$SMA) is quantified by measuring positively stained tissue area. 
\textcolor{marked}{The immunohistochemically stained areas were detected using the digital pathology software HALO 3.5 (Indica Labs; Albuquerque, NM, USA) with the Halo Area Quantification module. A threshold was applied to the resolved $\alpha$-SMA color component and used to determine its relative area.}
$\alpha$SMA is known to be increased in NAFLD \citep{Munsterman_Histopathology_2018}. Second, a pathologist's assessment of the severity of NAFLD using the NAFLD activity score (NAS) was performed \citep{Kleiner_Hepatology_2005}. When assigning the NAS, the pathologist combines three predefined histologic features known to change in NAFLD (steatosis, i.e., fat accumulation, inflammation, and ballooning, i.e., a form of hepatocyte cell death). In contrast to the methods above, the anomaly detector, which is trained on healthy tissue only, is neither tailored to NAFLD nor to any other specific disease.

\begin{figure*}[!htb]
\centering
\includegraphics[scale=1.0]{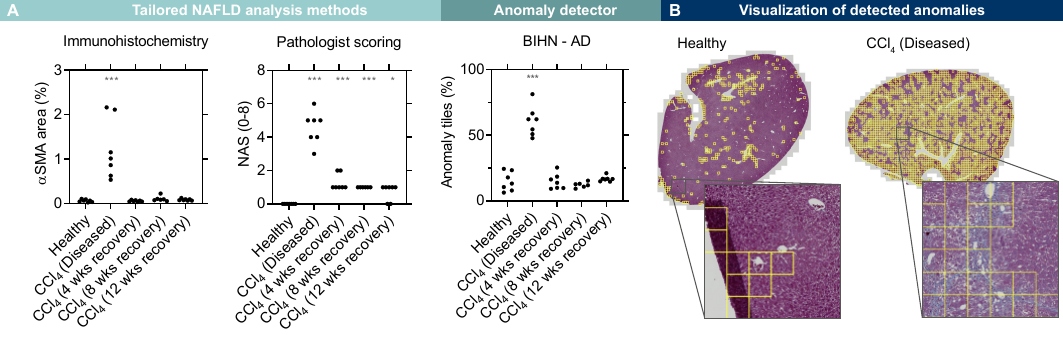}
\caption{\label{Fig:4} Detection of Non-Alcoholic
Fatty Liver Disease (NAFLD) in liver tissue. A: Comparison of the behavior of the BIHN based anomaly detection method with quantification methods specifically tailored to NAFLD, alpha Smooth Muscle Actin ($\alpha$SMA) immunohistochemical staining, and the NAFLD Activity Score (NAS) from pathologist examination. Each dot corresponds to a single WSI of the corresponding tissue type. Stars on the top of graphs show statistical significance of the change compared to the mean of control (healthy) group. B: Examples of detected anomalies in a control and a diseased liver.}
\end{figure*}

Fig.~\ref{Fig:4}A shows the results of the analysis of NAFLD pathology with the two tailored methods, IHC and pathologists scoring (NAS), compared to our anomaly detector.
Immunohistochemical analysis of $\alpha$SMA showed a significant increase in the area of $\alpha$SMA-positive liver tissue immediately after $CCl_4$ administration ($***, p \leq 0.001$, Mann-Whitney test, two-sided). No increased signal of $\alpha$SMA was observed in the recovery groups after 4, 8, and 12 weeks.
The NAS score also showed an increase in NAS immediately after $CCl_4$ administration ($p \leq 0.001$). Moreover, the NAS score remained slightly, but significantly increased in the recovery groups at 4 and 8 weeks ($***, p \leq 0.001$) and after 12 weeks ($*, p \leq 0.05$). This increase in NAS score after recovery was primarily caused by the liver inflammation component of the NAS.

For quantification of the response of our AD method, we computed the percentage of detected anomalous tiles. For liver tissue immediately after $CCl_4$ administration the AD method showed the same increase in signal ($p \leq 0.001$) as the targeted methods. In the recovery groups, the signal did not differ from that of the healthy control. Here, the pathologist could still detect a low level of inflammation as shown by the NAS score.
Even though our AD method was not developed for detection of specific pathological changes in the tissue, it closely repeats the response of the methods tailored to detection of NAFLD.

Fig.~\ref{Fig:4}B shows tissue examples from healthy (control) and $CCL_4$ groups, as well as detected abnormal tiles (yellow). In the $CCl_4$ treated liver, NAFLD features visible in the Masson-Trichrome stain include fibrosis (blue) and hepatocellular changes with micro and macro steatosis and ballooning as well as inflammatory infiltrates. The tiles with these tissue changes were correctly classified as abnormal. The false positive detections in the healthy liver include processing artifacts (folded tissue) and dilated blood vessels containing erythrocytes. Gathering additional training images of healthy tissue with such and a variety of other histological features can reduce false positives. 

\begin{figure*}[!htb]
\centering
\includegraphics[scale=1.0]{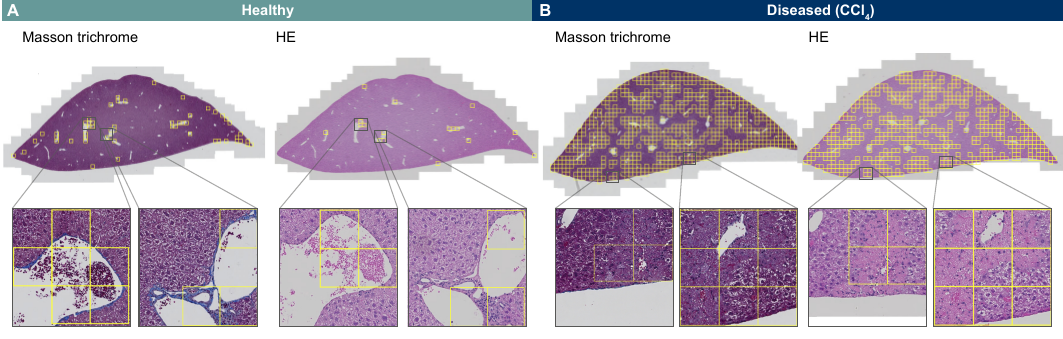}
\caption{\label{Fig:v2_6} \textcolor{marked}{Examples of anomaly detection in adjacent liver slices stained with Masson Trichrome and H\&E. A: Healthy tissue. False detections typically correspond to blood (left zoom in), vessels, localized inflammatory areas (right zoom in), or edges of the tissue. B: NAFLD tissue (CCL$_4$ induced). True detections correspond to necrotic tissue with some inflammatory cells.}}
\end{figure*}

\textcolor{marked}{Fig.~\ref{Fig:v2_6} gives additional examples of detections in WSI of healthy and diseased (NAFLD) tissue. For each of two cases we show adjacent tissue slices stained with Masson Trichrome and H\&E,  and apply anomaly detector that was trained on correspondingly stained tissue. One can see that detections are similarly localized in Masson Trichrome and H\&E stained tissue slices. Occasional false detections in healthy tissue  typically correspond to parts of vessels (white round areas), blood, or edges of the tissue, as can be seen in Fig.~\ref{Fig:v2_6}A. More diverse training datasets can probably reduce the number of such false detections. Widespread true detections in the diseased tissue in Fig.~\ref{Fig:v2_6}B correspond to necrotic areas with some inflammatory cells.}


\subsection{Detection of toxicological drug effects test}
\label{Sec:detection_adverse_events}


We used a pre-clinical toxicity study in order to verify if our method is capable of detecting pathological changes in \textcolor{marked}{H\&E stained} liver tissue previously diagnosed by an expert pathologist. In this study, mice received different doses of a substance in development for treating a particular disease (not liver related). The investigation by pathologists revealed that livers from treated mice showed dose-dependent hepatocellular cytoplasmic vacuolation. In addition, hepatocellular degeneration and necrosis with inflammation and mineralization as well as pigment deposition without a clear dose-relationship were seen in a significant number of treated animals from all dose groups.  

\begin{figure*}[!htb]
\centering
\includegraphics[scale=1.0]{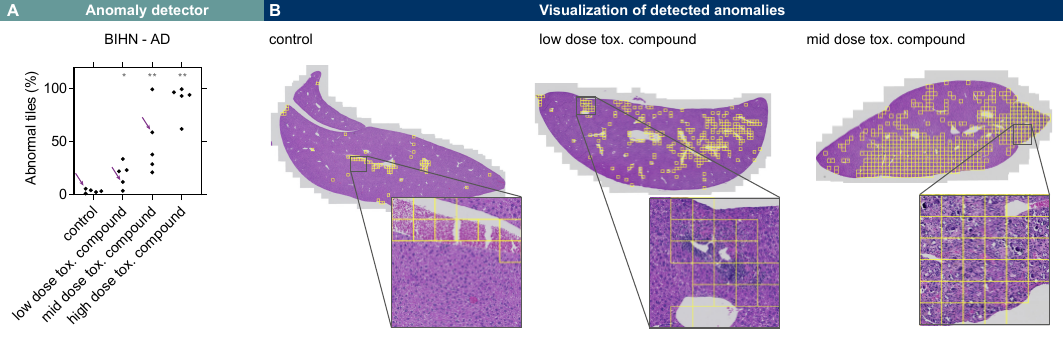}
\caption{\label{Fig:5} Detection of adverse drug reactions by the BIHN based anomaly detection. A: The developed AD method detects induced tissue alterations in the liver of mouse after administration an experimental compound. The fraction of abnormal tiles increases with the the dosage of the compound. The compound was previously found to have toxic side effects in toxicological screening by pathologists. Each dot corresponds to a single WSI. Three arrows correspond to three WSI examples given in B. Stars on the top of the graph show statistical significance of the change compared to the mean of control group. B: Examples of detected anomalies. In the control group (left image) blood and a few other not pathological structures result in a low level of false positives. Detections in compound treated groups (two right images) correspond to pathological alterations and were confirmed by a pathologist.}
\end{figure*}

Our method \textcolor{marked}{(BIHN model and one-class SVM were trained on H\&E training samples of healthy tissue)} reliably detected the highly variable abnormal morphological changes described above (Fig.~\ref{Fig:5}B, low dose: perivascular, inflammatory infiltrations and single cell necrosis; Fig.~\ref{Fig:5}B, mid dose: widespread necrosis with mineralization) and was able to discriminate them from the surrounding normal tissue. As can be seen in Fig.~\ref{Fig:5}A, the AD output is proportional to the studied substance dose and the severity of lesions damage as was diagnosed by the expert pathologist.  

\textcolor{marked}{The results underline the value of AD tools as a potential pre-screening modality, which can allow utilization of efficacy studies for screening not only the target organ (e.g., the lung), but also other organs possibly already showing toxicological side effects in the efficacy studies (e.g. in the liver or heart).} AD can be integrated in a digital pathology slide viewer to enable convenient pathologist review.    
Such AD tools may help in early detection of adverse findings and support the pathologists decision making. \textcolor{marked}{Benefits for drug discovery can be enormous, e.g., by early stopping of toxic compounds or compound classes, or by allowing early risk mitigation strategies for challenging drug targets. This could reduce costs in discovery research. The cost of the lead optimization and preclinical phases per new approved drug are estimated as 564 million\$ \citep{paul2010improve}. Since drug toxicity is the leading cause for preclinical early drug project termination \citep{waring2015analysis}, allowing earlier terminations can save substantial costs and result in a lower drug attrition in later phases. Moreover, there is an important ethical component, as AD may help reducing animal tests required for drug development following the 3R principles (i.e., reduction, refinement, replacement \citep{DOKE2015223}). If toxicity induced tissue alterations are discovered early, further in vivo efficacy tests and toxicological studies can be avoided.}

\subsection{Conclusion}

We developed an approach enabling detection of abnormal structures in histopathological images under the real-world environment. We introduced a \textcolor{marked}{recipe} for training image representations adapting them to the domain of histopathology, which allowed a standard anomaly detection to achieve strong performance. We collected and published a dataset of histopathological images of tissue with non-alcoholic fatty liver disease (NAFLD) that exhibit different abnormal variations and can be used as a benchmark for anomaly detection in histological images. Using this dataset we have shown that our approach outperforms the recent state-of-the-art and classical methods for anomaly detection.  
Moreover, we have shown that our method behaves similarly to established methods designed specifically to quantify NAFLD condition. We have also demonstrated that our approach was able to reveal tissue alteration in the liver resulted from side effects of a drug under development.
We can conclude that the developed approach is capable of detection of unknown adverse drug effects and has a potential of reducing attrition rates in the drug development.

\textcolor{marked}{Though our system is capable of reliably detecting conditions that typically spread over tissue, such as NAFLD, it might still be  challenging to detect localized abnormalities that may appear only within only a single or few tiles (e.g. focal inflammation). As was indicated in our ablation study in \ref{Sec:ablation_study}, higher complexity of the auxiliary task may further improve the performance, which might be required for reliable detection of small localized  anomalies. We, therefore, consider experimenting with an auxiliary segmentation task, e.g. segmentation of nuclei. Such a task will require manual annotation of ground truth masks or, alternatively, utilization of available systems for automated segmentation of particular histological structures. }

Currently, our system cannot use anomaly examples to improve its performance. Moreover, re-training one-class classifier on continuously gathered new normal data will not be efficient until the amount of the new data is relatively large. In future, we, therefore, aim at extending our anomaly detection approach to be able to gradually improve performance using both  anomalous and normal data that occasionally becomes available, i.e moving to continual learning regime. 

\section*{Acknowledgments}
We would like to thank Dr. Tanja Schönberger (Drug Discovery Sciences, Boehringer Ingelheim, Biberach, Germany) for help with the data curation and Dr. Martin Lenter (Drug Discovery Sciences, Boehringer Ingelheim, Biberach, Germany) for helpful discussions and project support. We would also like to thank Florian Rottach (IT EDP Central Data Sciences, Boehringer Ingelheim, Biberach, Germany) for his comments, which helped to improve the quality of the paper.


\bibliographystyle{model2-names.bst}\biboptions{authoryear}
\bibliography{refs}
\end{document}